\pdfoutput=1
\PassOptionsToPackage{dvipsnames}{xcolor}
\documentclass[11pt]{article}

\usepackage[]{acl}

\usepackage[utf8]{inputenc} 
\usepackage{times}
\usepackage{latexsym}
\usepackage{graphicx}
\usepackage{float}
\usepackage{subcaption}
\usepackage{soul}
\usepackage[most]{tcolorbox}
\usepackage{multirow}
\usepackage{array}
\usepackage{booktabs}
\usepackage{caption}
\usepackage{tabularx}
\usepackage{CJKutf8}
\usepackage{siunitx}

\definecolor{codebg}{gray}{0.95}  

\usepackage[T1]{fontenc}

\usepackage[utf8]{inputenc}

\usepackage{microtype}

\usepackage{inconsolata}

\usepackage{graphicx}

\title{A Multimodal, Multilingual, and Multidimensional Pipeline for Fine-grained Crowdsourcing Earthquake Damage Evaluation}

\author{
Zihui Ma$^{1}$, \quad
Lingyao Li$^{2}$\thanks{Corresponding authors.}, \quad
Juan Li$^{3}$, \\
\textbf{
Wenyue Hua$^{4}$, \quad
Jingxiao Liu$^{5}$, \quad
Qingyuan Feng$^{1}$, \quad
Yuki Miura$^{1}$\footnotemark[1]} \\
$^{1}$ New York University \quad
$^{2}$ University of South Florida \quad
$^{3}$ Google LLC \\
$^{4}$ University of California, Santa Barbara \quad
$^{5}$ Massachusetts Institute of Technology \\
\texttt{zihuima@nyu.edu, lingyao@usf.edu, leahlijuan1@gmail.com,} \\
\texttt{wenyue.hua@ucsb.edu, jingxiao@mit.edu, qf2093@nyu.edu, yuki.miura@nyu.edu}
}

\begin{document}
\maketitle
\begin{abstract} 
Rapid, fine-grained disaster damage assessment is essential for effective emergency response, yet remains challenging due to limited ground sensors and delays in official reporting. Social media provides a rich, real-time source of human-centric observations, but its multimodal and unstructured nature presents challenges for traditional analytical methods. In this study, we propose a structured Multimodal, Multilingual, and Multidimensional (3M) pipeline that leverages multimodal large language models (MLLMs) to assess disaster impacts. We evaluate three foundation models across two major earthquake events using both macro- and micro-level analyses. Results show that MLLMs effectively integrate image-text signals and demonstrate a strong correlation with ground-truth seismic data. However, performance varies with language, epicentral distance, and input modality. This work highlights the potential of MLLMs for disaster assessment and provides a foundation for future research in applying MLLMs to real-time crisis contexts. The code and data are released at: \url{https://github.com/missa7481/EMNLP25_earthquake}
\end{abstract}

\section{Introduction}

Efficient and comprehensive disaster damage assessment is critical for informing emergency operations and disaster relief \cite{ma2024surveying, shan2019disaster,miura2021methodological}. Conventional techniques such as hazard models, expert inspections, and ground-based instruments have supported the characterization of post-disaster conditions \cite{butenuth2011infrastructure, torok2014image, tate2015uncertainty}. Recently, social media crowdsourcing has emerged as an additional source of information \cite{kryvasheyeu2016rapid, ma2024surveying}, offering large volumes, near-real-time insights from those affected communities \cite{li2021data, ma2024investigating}. More importantly, social media offers passive human observations, often capturing nuanced perspectives such as emotional reactions, indoor damage, and first-hand observations \cite{ma2024surveying, li2023exploring}. These human-centric signals add a layer of damage representation to the conventional methods. 


However, earlier machine learning methods frequently relied on hand-crafted features and domain-specific models, which required significant manual effort to extract structured insight \cite{devaraj2020machine, o2020deep, ma2024surveying}. Moreover, they 
often lack the generalizability to apply across multiple disasters occurring in different locations with different languages, or involving varying damage levels, as models trained on one dataset (\emph{e.g.}, data from a specific disaster or spoken language) may not perform well on another. Additionally, diverse multimodal inputs pose challenges for analysis.
Recent advances in foundation MLLMs have demonstrated potential for cross-modal and multilingual understanding across diverse data sources. 
Though promising, 
it is unclear whether MLLMs can support fine-grained damage assessment, including structural and environmental impacts, interior damage, and human experiences across different language regions. Moreover, their scalability and generalizability across disasters and geographies have not been systematically evaluated, as this could be critical for supporting disaster managers in implementing quick disaster relief. 

To address these gaps,
we propose a structured ``Multimodal, Multilingual, and Multidimensional'' (3M) pipeline integrating data collection, multimodal damage classification, and model evaluation. Our pipeline relies on the reasoning abilities of MLLM to extract interpretable, event-relevant insights from large-scale social media streams. We evaluate this pipeline using two sudden-onset earthquake events: the 2019 Ridgecrest earthquake in California and the 2021 Fukushima earthquake in Japan. Across these two case studies, we assess three top-performing foundation models, including Gemini-2.5-Flash (hereafter \textit{Gemini}) \cite{team2023gemini}, LLaVA 3-8B (hereafter \textit{LLaVA}) \cite{intel_llava_llama3_8b}, and Qwen 2.5-VL-7B (hereafter \textit {Qwen}) \cite{qwen2.5}, to explore their ability to understand multilingual content, reason across modalities, and generate consistent damage-level predictions.
The study aims to answer the following questions through macro- and micro-perceptions:

\begin{itemize}
    \item Can MLLMs provide reliable and fine-grained damage assessments of textual and image information posted on social media after disasters?
    \item To what extent do MLLMs generalize across disaster contexts, with respect to factors such as input modality and prompt sensitivity?
\end{itemize}

Our findings suggest that MLLMs exhibit strong capabilities in event localization, image-text fusion, and perceptual damage estimation. The models correlate near-moderate positive (r=\textasciitilde 0.5) to high (r=0.78) with ground-truth seismic intensity data and demonstrate interpretable reasoning patterns. However, we also observe variations in performance depending on 
linguistic context and event proximity. These findings highlight both the promise and limitations of current LLMs, and point toward future directions for model adaptation and disaster-specific fine-tuning.

\section{Related Work}

\subsection{Earthquake Damage Assessment}
\par
Recent advances in earthquake damage assessment span physics-based models, machine learning, and new sensing modalities, each balancing trade-offs in accuracy, scalability, and timeliness. 
Traditional approaches, such as FEMA’s HAZUS and P-58 frameworks~\cite{schneider2006hazus,hamburger2012fema}, rely on structural mechanics to estimate probabilistic damage and losses. 
While interpretable and robust, these methods are computationally demanding, depend on expert input, and often lack the spatial resolution and speed needed for rapid, localized assessments. 
Their reliance on coarse, regional building inventories and categorical outputs (e.g., “moderate” or “extensive” damage) limits their utility in dynamic, real-world disaster response. Moreover, their reliance on physical instrumentation limits deployment coverage and often excludes human-centered perspectives on impact.

\par

\par
Building on machine learning advances, researchers have begun exploring novel data sources, such as crowdsourced social media. 
Existing literature has used sentiment analysis \cite{li2025mining, myint2024unveiling, amangeldi2024understanding, subbaiah2024efficient}, topic modeling \cite{ma2024investigating,mihunov2022disaster, mehmood2024named}, and text classification  \cite{xie2022multi, yin2024crisissense, garcia2025explainable} to support hazard monitoring, communication, damage assessment, and behavioral analysis \cite{ma2024surveying}. 
Yet despite their promise, these sources are often used in isolation. 
Most existing frameworks do not integrate these diverse inputs into a unified pipeline. 
They are commonly limited to a single data type (text or image), rely heavily on English-language content, and lack systematic incorporation of damage granularity aligned with MMI levels. 
This leads to a fragmented understanding of earthquake impacts, with missed opportunities for timely, contextualized, and community-aware responses.

\subsection{Multi-modal LLMs Applications}
Multimodal foundation models have emerged as powerful tools for integrating diverse data types, revolutionizing capabilities across scientific domains. Models such as GPT-4V \cite{wu2023early}, Gemini \cite{team2023gemini}, and Claude 3 \cite{kevian2024capabilities} are capable of understanding and reasoning over multimodal data, including text, images, video, and numerical data, demonstrating remarkable performance in tasks requiring cross-modal understanding. These models have shown effectiveness in analyzing complex scientific imagery alongside textual annotations, enabling new approaches to data fusion in fields ranging from bioinformatics \cite{luo2024biomedgpt, wang2025large, liu2024geneverse} to astronomy \cite{rizhko2024self, mishra2024paperclip}. 

The application of multimodal foundation models has expanded beyond traditional scientific domains to critical social applications, particularly in disaster response \cite{hughes2025seeing, odubola2025ai, lei2025harnessing} and social media analysis \cite{thapa2025large, de2023llm}. These models are leveraged to interpret structural damage by aerial imagery \cite{jiang2025eaglevision} and social media post analysis \cite{sharma2024experts} to prioritize emergency response resources \cite{yu2024multimodal}, using both visual and textual contents to achieve a nuanced understanding of real-time information during crisis events. 

\section{3M Pipeline}

\begin{figure*}[htbp]
    \centering
    \includegraphics[width=0.8\textwidth]{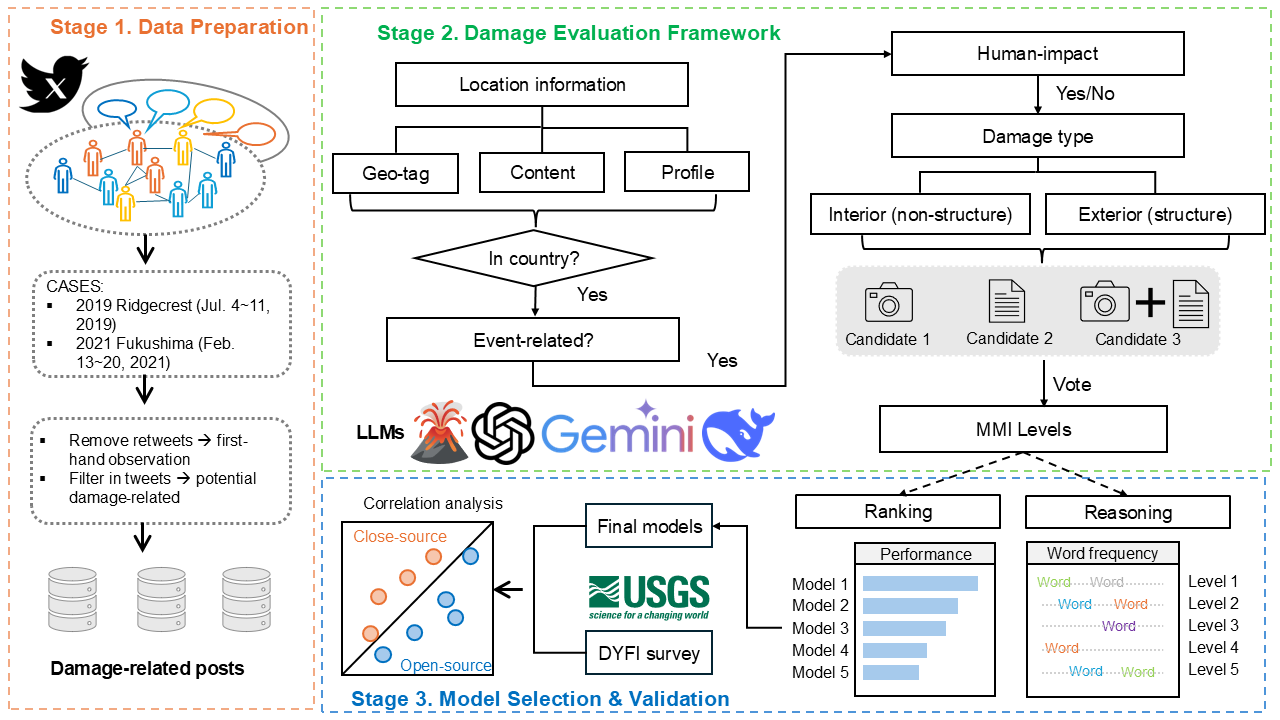}
    \caption{Proposed 3M pipeline, which integrates data preparation, damage evaluation framework, and model selection and validation for social media–based earthquake assessment.}
    \label{fig: framework}
\end{figure*}

To achieve fine-grained earthquake damage assessment from social media, we develop the 3M pipeline, illustrated in Figure \ref{fig: framework}. The pipeline consists of three primary stages, and each component is detailed in the following subsections. 
\paragraph{Data Preparation}

Twitter (now rebranded as X) is a microblogging and social networking platform that allows users to share short messages known as ``tweets.'' Since the data in this study are collected prior to the rebranding, we refer to the platform as ``Twitter'' and use the term ``tweets'' for consistency. This study focuses on two representative earthquake events: (1) the 2019 Ridgecrest earthquake in California and (2) the 2021 Fukushima earthquake in Japan. These cases are selected because they occurred in seismically active regions with established disaster response systems. 

Then, tweets are collected using the Twitter Search API in ``near-real-time'' with the keyword ``earthquake.'' For the Ridgecrest event, tweets are collected from July 4 to 10, 2019; for the Fukushima event, from February 13 to 17, 2021. 
Following the compilation of the initial dataset, a filtering process is applied to identify tweets containing damage-related content. Guided by prior research \cite{li2023exploring}, we construct a library of filter terms (\emph{e.g.}, ``damage,'' ``injury,'' ``hurt,'' ``die,'' ``kill''), accounting for common word variants (\emph{e.g.}, ``damage,'' ``damages,'' ``damaged''). This filtering yields a refined dataset, referred to as the ``damage-related dataset.'' After applying these criteria, the final dataset consists of 41,431 damage-related tweets for the 2019 Ridgecrest earthquake and 49,539 for the 2021 Fukushima earthquake, which are used for the subsequent analysis. The full list of filter terms is provided in the Appendix.

\paragraph{Damage Evaluation Framework}
The evaluation of earthquake damage through social media content necessitates a structured and multi-stage analytical framework. For any given Twitter post, the assessment initially establishes event relevance through a two-fold verification process. First, spatial contextualization is conducted using a tiered approach that incorporates (1) geotag metadata, (2) content-based geographic references, and (3) user profile registration information. Among these, we prioritize geotagged metadata, which provides the most precise spatial signal \cite{stock2018mining,doran2014accurate}. When geotag data is unavailable or ambiguous, we rely on content-based inference (\emph{e.g.}, mentions of place names or landmarks) and, subsequently, on user profile location. In cases where multiple geographic scales are mentioned (\emph{e.g.}, city and neighborhood), the framework returns to the most granular location available. Second, the framework verifies the targeted seismic event to ensure analytical specificity.

Upon confirmation of relevance, the damage assessment protocol follows a hierarchical classification approach. The primary analysis differentiates between human-impact scenarios and non-human structural consequences. This bifurcation enables specialized examination of non-human impacts, which are further categorized into interior non-structural damage (\emph{e.g.}, cracked interior walls, broken windows) and exterior structural damage (\emph{e.g.}, building façade collapse, fallen infrastructure). It employs MLLMs to synthesize both textual narratives and visual documentation from social media posts. Based on the aggregated damage indicators, each post is assigned a Modified Mercalli Intensity (MMI) level. The MMI scale is a qualitative, ten-point system that characterizes earthquake intensity based on human perception and observable environmental and structural effects. Unlike instrumental magnitude scales, MMI provides a human-centered measure of impact, making it a widely adopted standard in post-earthquake reporting and risk communication. Detailed descriptors of the MMI scale used in this study are provided in the Appendix \ref{MMI}. The use of MMI levels enables standardized comparisons of seismic impacts across geographic regions and disaster events. We leverage few-shot \cite{brown2020language} chain-of-thought (CoT) \cite{wei2022chain} prompting for model evaluation.


\paragraph{Model Selection and Validation}
This stage involves both quantitative and interpretive evaluation. We evaluate eight state-of-the-art multimodal foundation models, including leading commercial and open-source systems: GPT-4.1, GPT-4.1-mini, GPT-4.1-nano, GPT-4o, GPT-4o-mini, Gemini-2.5-Flash, LLaVA 3–8B, and Qwen 2.5-VL-7B. These models are selected based on their reported performance in vision-language tasks and their accessibility for benchmarking \cite{wang2024picture, guruprasad2024benchmarking}.

Using a randomly selected sample of damage-related tweets, each model generated MMI levels through the previous stage. Human-labeled ground-truth classes are based on the agreement of two independent annotators using the same damage framework. Pearson correlation scores are used to rank each model’s performance in terms of alignment with official seismic intensity data. 
The full comparative results 
are provided in the Appendix \ref{model_compartion}. Based on this analysis, Gemini, LLaVA, and Qwen were selected for following analysis, considering computational efficiency and practical deployment constraints. To assess the overall accuracy, we perform a correlation analysis between the model-generated MMI levels and ground-truth labels derived from the USGS "Did You Feel It?" (DYFI) survey \cite{wald2011usgs}, a crowdsourced platform that collects public reports of perceived shaking intensity following an earthquake.

Following quantitative validation, we further investigate the reasoning transparency of the top-performing models to understand how MLLMs estimate MMI levels. Specifically, we analyze the textual justifications generated by each model, focusing on the lexical features that underlie their classification decisions. To this end, we conduct a unigram-level TF-IDF analysis to identify high-weighted terms associated with different MMI levels. This analysis reveals the most influential words contributing to the model’s classification decision. By analyzing the alignment between high-weighted terms and relevant damage descriptors, we assess whether a model's internal logic aligns with human-interpretable features. 

\begin{figure*}[!ht]
 \centering
 \includegraphics[width=\textwidth]{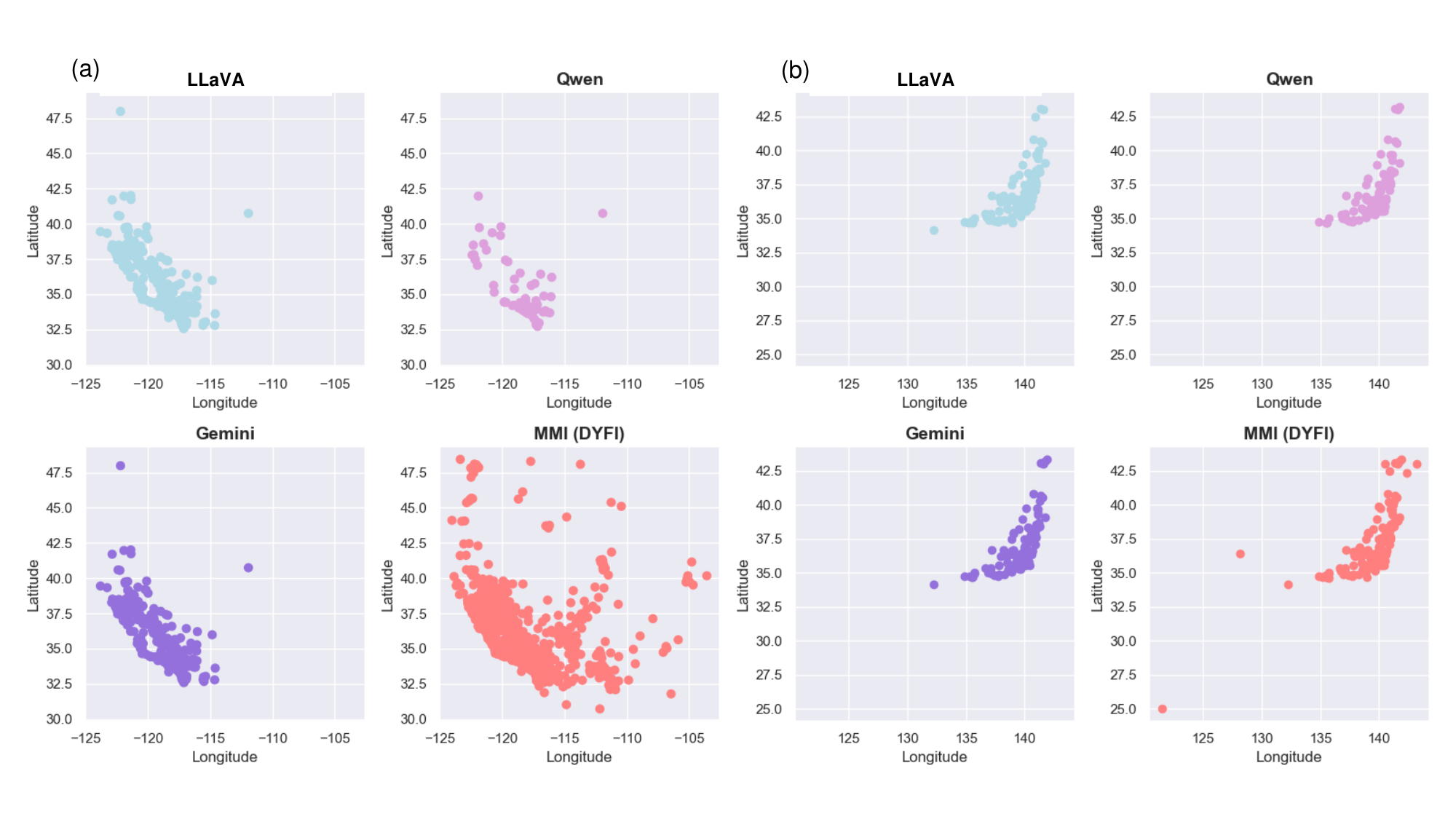}
 \vspace{-25pt}
 \caption{Spatial distribution of (a) Ridgecrest and (b) Fukushima data points identified by LLaVA, Qwen, and Gemini compared to DYFI MMI reports.}
 \vspace{-5pt}
 \label{fig:cor_spatial_all}
\end{figure*}

\section{Experiments and Results}
In this section, we present the main experimental results and analysis. The first part focuses on macro-level evaluation at the pipeline level, including two earthquake case studies and an assessment of epicentral distance effects on model performance. The second part provides a micro-level analysis at the model level, examining impact of input modality, model prompt sensitivity, and detailed analysis of MLLM reliability based on CoT outputs.

\subsection{Macro-level Analysis}
\paragraph{2019 Ridgecrest Earthquake}
Figure \ref{fig:cor_spatial_all}(a) shows the spatial distributions of social media-derived locations identified by three selected MLLMs in comparison to DYFI MMI scales.
Overall, the results suggest that the models are capable of extracting relevant location and event information from tweets, as evidenced by the clustering of identified points near the earthquake epicenter (35.766°N 117.605°W). 
Qwen demonstrates relatively weak performance in spatial coverage, with fewer identified points and reduced geographic spread. This may be due to its pretraining focus on Chinese-language data.

\begin{figure*}[ht]
   \centering
   \includegraphics[width=\textwidth]{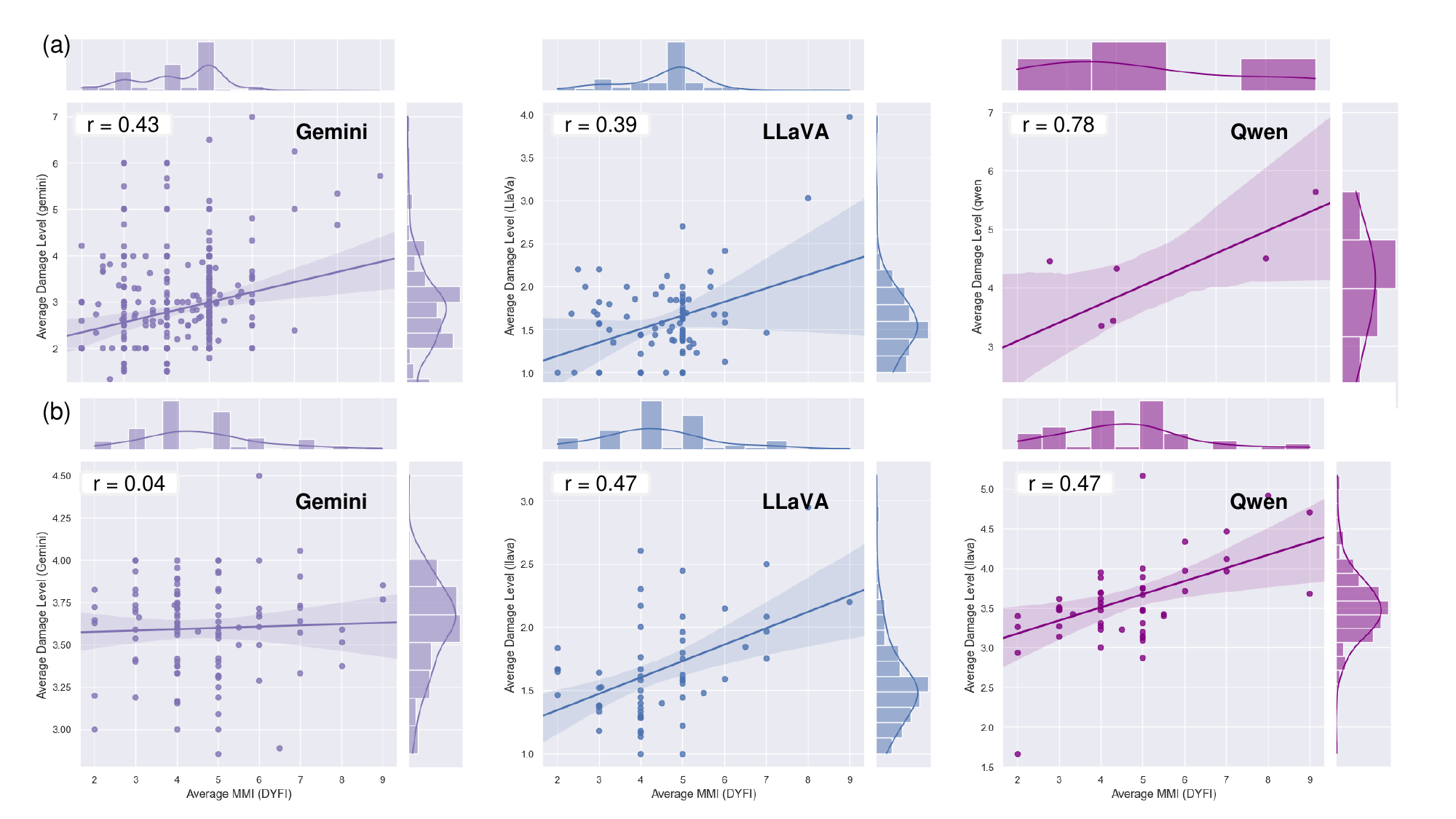}
   \vspace{-15pt}
   \caption{Correlation between model-estimated average damage levels and DYFI MMI levels for (a) Ridgecrest and (b) Fukushima earthquakes.}
   \vspace{-5pt}
   \label{fig:cor_model}
\end{figure*}

We further assess the models' ability to infer earthquake damage levels. 
Figure \ref{fig:cor_model}(a) presents the city-level correlations between model-estimated average damage levels and DYFI MMI data. All models show near-moderate to high positive agreements, as measured by Pearson correlation coefficients. Interestingly, Qwen achieves the highest correlation (r = 0.78), suggesting that although its spatial recall is limited, it may still be effective at identifying intensity-related cues from text and imagery. 


\paragraph{2021 Fukushima Earthquake}
Similarly, we apply 3M pipeline to the 2021 Fukushima Earthquake in Japan with predominantly Japanese social media content. 

Most of the identified data points cluster near the earthquake epicenter (37.730°N, 141.595°E), and their spatial distributions align closely with the DYFI MMI data (Figure \ref{fig:cor_spatial_all} (b)). All three models capture nearly the full range of earthquake-affected locations. 
Their performance diverges when it comes to fine-grained damage level assessment. As shown in Figure \ref{fig:cor_model} (b), Gemini exhibits a weak correlation between model-inferred damage levels and DYFI MMI scores (r = 0.04), 
 In contrast, LLaVA and Qwen achieve near-moderate correlations (r = 0.47 for both), reflecting a better understanding of MMI-scale damage in Japanese content. Although the overall correlation values for LLaVA and Qwen are similar, their strengths differ by intensity range. Qwen demonstrates a more precise differentiation between MMI levels 3 and 4, indicating sensitivity to moderate damage. LLaVA, on the other hand, performs more reliably in the lower MMI range (levels 1 to 3). 

\paragraph{Epicenter Distance}   

We examine 
the correlation between estimated MMI levels and epicentral distance to assess the spatial sensitivity of model estimations, using results from the best-performing models for each case: Gemini for the Ridgecrest event and LLaVA for the Fukushima event (Figure \ref{fig:distance}). 
In both cases, a negative correlation was observed, consistent with the principle of seismic attenuation, where shaking intensity typically decreases with increasing distance from the epicenter. The trend was stronger in the English-language Ridgecrest case, suggesting language familiarity may influence a model’s ability to learn physically grounded patterns. Notably, Gemini identified a concentration of high-MMI predictions within ~200 km of the epicenter, especially in densely populated areas (\emph{e.g.}, Los Angeles), indicating its ability to focus on high-risk urban zones. 

\subsection{Micro-level Analysis}
\begin{figure}[!ht]
  \centering
  \includegraphics[trim=20 30 10 20, clip,width=\columnwidth]{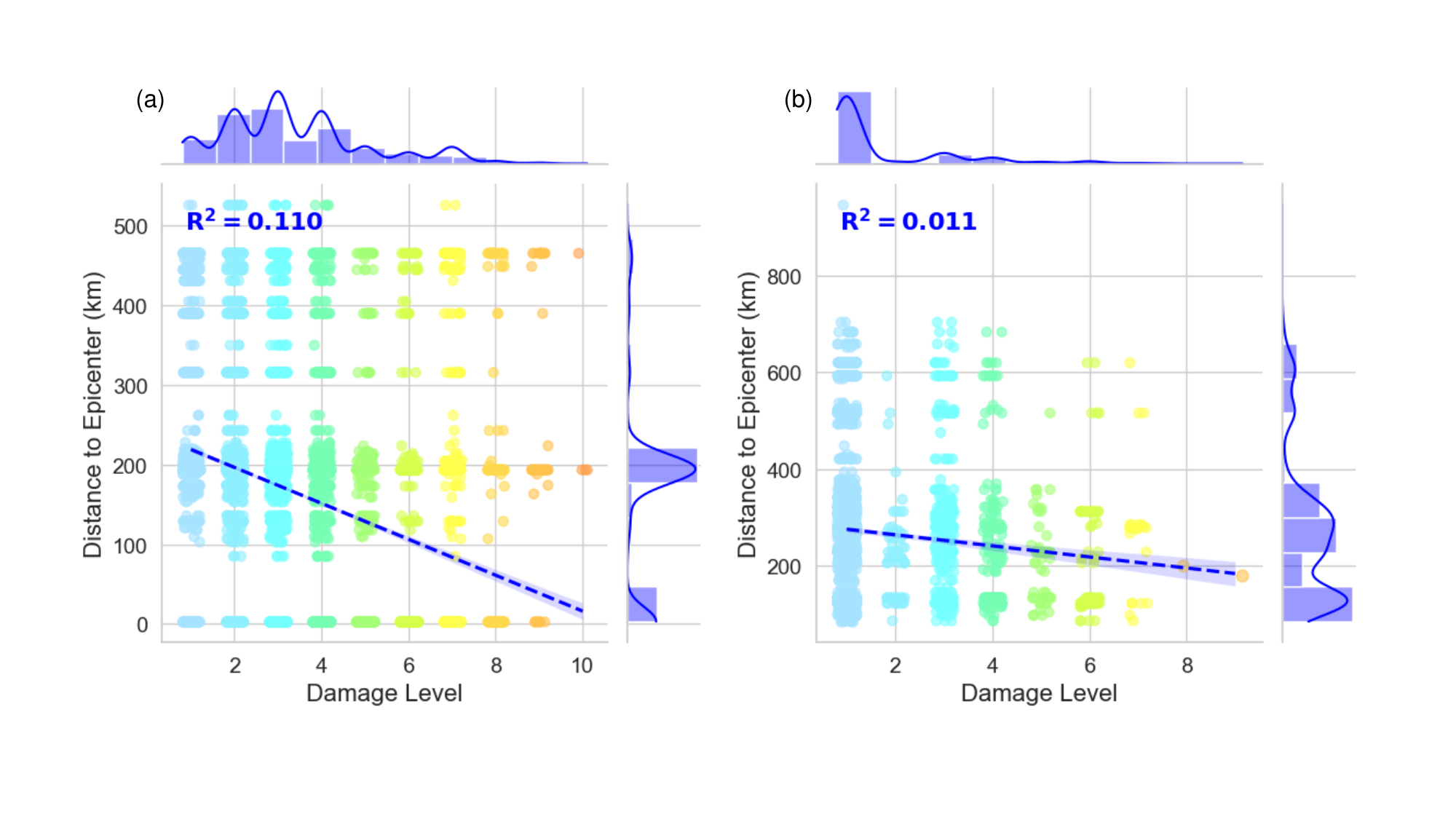}
  \caption{Epicentral distance vs. model estimated MMI values for the (a) Ridgecrest earthquake;(b) Fukushima earthquake.}
  \label{fig:distance}
\includegraphics[trim=20 120 10 30, clip, width=\columnwidth]{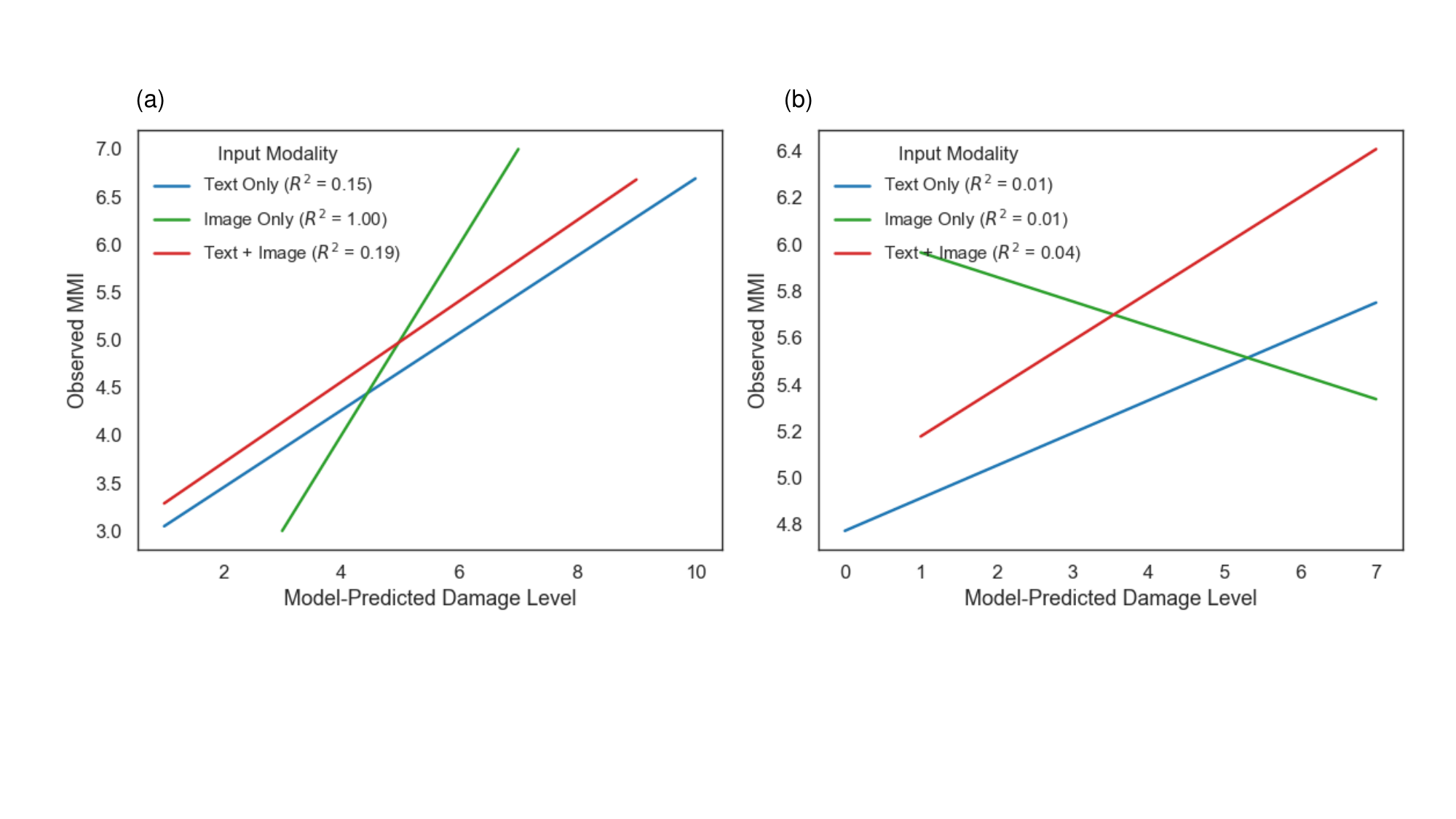}
\caption{Correlation between model-estimated and DYFI MMI levels across input types for (a) Ridgecrest and (b) Fukushima earthquakes.}
  \label{fig:data_type}
\end{figure}

\paragraph{Input Modality.} The choice of input modality directly influences the framework's evaluation performance. While social media platforms are primarily text-driven, the effectiveness of visual information and its combination with text for damage assessment remains underexplored. Thus, we evaluate model performance across three input configurations: text-only, image-only, and text-image fusion, as implemented in 3M pipeline. Correlation analysis between predicted and DYFI MMI levels across these settings is shown in Figure \ref{fig:data_type}. In both earthquake cases, models fusing textual and visual content strongly correlated with observed MMI, reinforcing prior findings in multimodal literature that show the benefit of cross-modal integration \cite{merlo2010cross, maragos2008cross, wang2025cross}. Conversely, models relying solely on visual inputs show diminished performance, particularly in the non-English Fukushima dataset, where image-only analysis 
was often based on non-damage-related content visuals, such as selfies, emojis, or screenshots, which lacked direct evidence of structural damage or event relevance. 


\paragraph{Prompt Sensitivity}  
Given that variations in prompt phrasing could impact model performance, 
it is crucial to evaluate the sensitivity of MLLMs to different prompt formulations \cite{sclar2023quantifying, zhuo2024prosa, chatterjee2024posix}. This section builds on our earlier results by examining whether slight variations in prompts affect the models’ outputs. 
To explore this, we randomly selected 50 tweets. 
For each tweet, we created seven paraphrased versions of the original prompt using GPT-4o. These paraphrases reword the instructions while keeping the meaning the same. All rewritten prompts were manually checked to ensure clarity and correctness. The full list of prompt versions is provided in the Appendix~\ref{prompt_versions}. We analyzed the impact of prompt variation across four output types: damage level, confidence score, and categorical judgments such as damage type and human impact. For numerical outputs, we used mean and standard deviation to measure variability. For categorical outputs, we used Cramér’s V \cite{cramer1999mathematical} (Equation \ref{eq:cramers_v}) to measure how often the predictions changed across prompts, where values closer to 0 mean low sensitivity, and values closer to 1 mean high sensitivity.

\begin{equation}
V = \sqrt{ \frac{\chi^2}{n \cdot (k - 1)} }
\label{eq:cramers_v}
\end{equation}

\noindent
where (1) $V$ is Cramér’s V coefficient (2) $\chi^2$ is the chi-squared statistic derived from the contingency table (3) $n$ is the total number of observations, and (4) $k$ is the number of categories in the smaller of the two variables, \emph{i.e.}, $\min(\text{rows}, \text{columns})$.

\begin{table*}[htbp]
\centering
\tiny
\caption{Cramér’s V scores for human impact and damage type across prompt versions and models}
\begin{tabular}{llcccccc}
\toprule
\multirow{2}{*}{\textbf{Event}} & \multirow{2}{*}{\textbf{Prompt}} &
\multicolumn{2}{c}{\textbf{Gemini}} &
\multicolumn{2}{c}{\textbf{Qwen}} &
\multicolumn{2}{c}{\textbf{LLaVA}} \\
\cmidrule(lr){3-4} \cmidrule(lr){5-6} \cmidrule(lr){7-8}
& & \textbf{Human Impact} & \textbf{Damage Type} & \textbf{Human Impact} & \textbf{Damage Type} & \textbf{Human Impact} & \textbf{Damage Type} \\
\midrule

\multirow{1}{*}{\shortstack[l]{2019 Ridgecrest earthquake}} 
& v1-v7 & 0.170 & 0.225 & 0.218 & 0.464 & 0.636 & 0.511 \\

\midrule

\multirow{1}{*}{\shortstack[l]{2022 Fukushima earthquake}} 
& v1-v7 & 0.502 & \textcolor{red}{0.771} & 0.224 & 0.624 & 0.578 & 0.587 \\

\bottomrule
\end{tabular}
\label{tab:cramersv_results}
\end{table*}

\begin{table*}[htbp]
\centering
\tiny 
\caption{Damage level and confidence scores across prompt versions and models}
\begin{tabularx}{\textwidth}{ll
  *{4}{>{\centering\arraybackslash}X}  
  *{4}{>{\centering\arraybackslash}X}  
  *{4}{>{\centering\arraybackslash}X}  
}
\toprule
\multirow{2}{*}{\textbf{Event}} & \multirow{2}{*}{\textbf{Prompt}} &
\multicolumn{4}{c}{\textbf{Gemini}} &
\multicolumn{4}{c}{\textbf{Qwen}} &
\multicolumn{4}{c}{\textbf{LLaVA}} \\
\cmidrule(lr){3-6} \cmidrule(lr){7-10} \cmidrule(lr){11-14}
& & DL\_mean & DL\_std & Conf\_mean & Conf\_std
  & DL\_mean & DL\_std & Conf\_mean & Conf\_std
  & DL\_mean & DL\_std & Conf\_mean & Conf\_std \\
\midrule

\multirow{7}{*}{\shortstack[l]{2019 Ridgecrest\\earthquake}} 
& v1 & 3.333 & 1.325 & 0.786 & 0.073 & 3.600 & 1.694 & 0.850 & 0.059 & 1.867 & 1.548 & 0.853 & 0.090 \\
& v2 & 2.795 & 1.490 & 0.847 & 0.098 & 2.588 & 2.311 & 0.887 & 0.054 & 0.769 & 1.945 & 0.915 & 0.141 \\
& v3 & 2.317 & 1.572 & 0.906 & 0.101 & 2.606 & 2.304 & 0.911 & 0.066 & 0.769 & 2.026 & 0.962 & 0.070 \\
& v4 & 2.683 & 1.980 & 0.894 & 0.094 & 3.100 & 1.919 & 0.883 & 0.091 & 1.867 & 1.388 & 0.915 & 0.097 \\
& v5 & 3.095 & 1.923 & 0.875 & 0.091 & 3.152 & 2.224 & 0.809 & 0.109 & 2.104 & 2.479 & 0.840 & 0.184 \\
& v6 & 2.683 & 1.559 & 0.848 & 0.082 & 3.212 & 2.162 & 0.800 & 0.080 & 1.940 & 0.752 & 0.876 & 0.072 \\
& v7 & 1.930 & 1.486 & 0.900 & 0.105 & 4.188 & 2.583 & 0.883 & 0.066 & 2.720 & 2.777 & 0.832 & 0.118 \\
\midrule

\multirow{7}{*}{\shortstack[l]{2022 Fukushima\\earthquake}} 
& v1 & 3.838 & 1.041 & 0.722 & 0.062 & 3.222 & 1.502 & 0.850 & 0.069 & 3.333 & 1.325 & 0.786 & 0.073 \\
& v2 & 3.563 & 1.105 & 0.731 & 0.098 & 2.667 & 2.075 & 0.837 & 0.099 & 3.167 & 1.324 & 0.782 & 0.100 \\
& v3 & 2.333 & 1.875 & 0.942 & 0.063 & 2.444 & 2.470 & 0.815 & 0.151 & 3.167 & 1.998 & 0.915 & 0.060 \\
& v4 & 4.000 & 1.732 & 0.797 & 0.077 & 3.124 & 2.378 & 0.805 & 0.076 & 2.625 & 2.042 & 0.863 & 0.078 \\
& v5 & 4.630 & 1.245 & 0.687 & 0.071 & 2.667 & 1.637 & 0.873 & 0.070 & 2.167 & 2.681 & 0.633 & 0.334 \\
& v6 & 2.214 & 1.528 & 0.786 & 0.063 & 3.955 & 2.645 & 0.763 & 0.127 & 0.167 & 0.817 & 0.852 & 0.260 \\
& v7 & 1.243 & 0.760 & 0.801 & 0.092 & 4.239 & 2.508 & 0.787 & 0.239 & 1.000 & 2.703 & 0.850 & 0.269 \\
\bottomrule
\end{tabularx}
\label{tab:damage_confidence_summary}
\end{table*}
As shown in Table \ref{tab:cramersv_results} and Table \ref{tab:damage_confidence_summary}, models evaluated on the Fukushima dataset exhibit greater sensitivity 
than those tested on Ridgecrest data. This pattern was particularly pronounced for Gemini, which demonstrates substantial response variability when processing Japanese-language tweets. Conversely, Qwen displays the most stable performance, showing minimal variation in damage level assessments and confidence scores, though it exhibits greater inconsistency in damage type classification. 

Despite variations in categorical classifications, most models maintain relatively stable MMI level predictions, with standard deviations typically ranging between 1 and 2. This indicates that while prompt formulation can influence specific classification details, overall assessments remain reasonably consistent. A similar pattern is observed in confidence scores, suggesting that models maintain comparable levels of certainty regardless of instructional phrasing.


\paragraph{Reasoning Reliability Evaluation} 

\begin{figure*}[!ht]
 \centering
 \includegraphics[width=\textwidth]{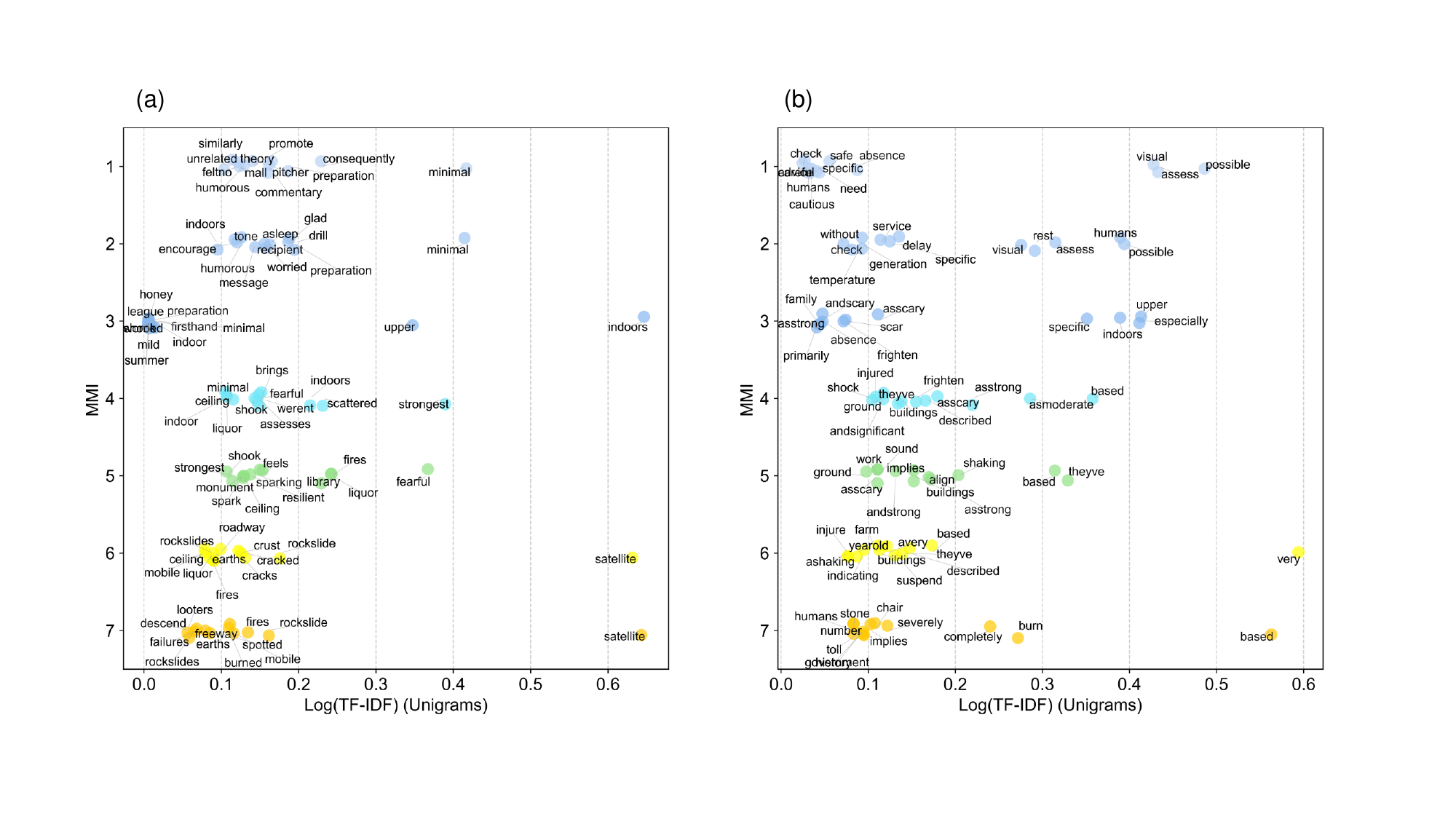}
 \vspace{-5pt}
 \caption{Reasoning reliability evaluation for (a) Ridgecrest and (b) Fukushima.}
 \vspace{-5pt}
 \label{fig:reasoning_all}
\end{figure*}

To better understand how models arrive at their predictions, we conduct an analysis of the language used in their free-text justifications for estimated MMI levels, presenting a taxonomy of lexical patterns associated with different intensity levels. 
For the Ridgecrest earthquake (Figure \ref{fig:reasoning_all} (a)), Gemini exhibits a progression in reasoning. At lower MMI levels (0–3), the model frequently uses terms such as “minimal,” “preparation,” “indoors,” and “worries,” suggesting a focus on psychological response and perceived safety. As the MMI increases to moderate levels (4–5), emotionally charged terms like “shock” and “fearful” become more common. At higher intensity levels (6–9), the model increasingly references concrete environmental and structural cues, using terms like “rockslides,” “cracked,” and “roadway.” It later shifts toward cascading impact language with words like “fires” and “burned.”


For the Fukushima earthquake (Figure \ref{fig:reasoning_all}(b)), LLaVA centers on perceived safety and emotional state, with terms like “visual,” “safe,” and “scary” at a lower MMI level. At moderate to higher MMI levels, the model references physical objects with increasing specificity, such as terms of “building,” “ground,” and “chair.” For severe impacts, the model incorporates stronger terms such as “injured,” “suspend,” and “severely.”  Interestingly, LLaVA often uses hedging terms (e.g., ‘possible’, ‘indicating’), suggesting a more cautious or probabilistic reasoning style.



\section{Discussion}
\textbf{Can MLLMs provide reliable and fine-grained damage assessments using multilingual textual and image information posted on social media after disasters?}
Our experimental results demonstrate that state-of-the-art MLLMs possess substantial potential for fine-grained earthquake damage assessment. 
The effectiveness of the 3M pipeline across both English- and non-English-language contexts further demonstrates the multilingual capabilities of MLLMs. With appropriate language-aligned foundation models, the pipeline can be generalized to additional languages and extended to other disaster types (\emph{e.g.}, wildfires, hurricanes) through prompt adaptations. This flexibility underscores the scalability of our approach across geographic and hazard domains.

Despite promising results, we observed some model-level performance variation. Qwen demonstrated the most consistent performance across languages, making it suitable for multilingual contexts, while Gemini and LLaVA excelled in urban, English-dominant settings. All models were more reliable at low to moderate damage levels, with reduced accuracy at higher intensities. It is likely due to training and data sparsity. Additionally, model estimation were influenced by epicentral distance, with better performance in densely populated urban areas. This pattern suggests that MLLMs capture attenuation effects but are also shaped by spatial disparities in social media activity. For real-world applications, decision-makers should account for these biases and consider complementary data sources or localized calibration when applying the 3M pipeline beyond densely populated regions.

\textbf{To what extent do MLLMs generalize across disaster contexts, with respect to factors, such as input modality and prompt sensitivity?}
Our micro-level analysis further guides for the deployment of MLLMs in disaster contexts. First, modality analysis confirms that multimodal input fusion improve both accuracy and robustness in damage classification. We recommend extending this approach to include cross-modal fusion of additional modalities such as video, audio, and geospatial data (\emph{e.g.}, satellite imagery, street-level views). Second, prompt sensitivity evaluation reveals that current MLLMs exhibit variability in multilingual contexts, especially in response to subtle changes in instruction phrasing. While categorized classification outputs (\emph{e.g.}, damage type, human impact) are relatively stable, inconsistencies may arise in edge cases. We recommend prompt standardization, pre-deployment testing, and ensemble prompting strategies to reduce sensitivity in multilingual or low-resource environments. Lastly, our reasoning analysis highlights differences in model interpretability and internal logic. For example, Gemini shifts from emotional to structural and cascading-impact cues as damage severity increases, while LLaVA adopts a more visually grounded but cautious reasoning style. These patterns suggest that decision-makers should consider not only performance metrics but also reasoning transparency and alignment with operational needs when selecting models for deployment.

\section{Conclusion}
This study introduces a structured 3M pipeline for social media–based earthquake damage assessment. The pipeline systematically integrates data preparation, multimodal classification, and model evaluation, providing a scalable framework for rapid and fine-grained disaster analysis. Applied to two real-world earthquake events, the pipeline demonstrates its effectiveness across languages, geographies, and damage dimensions. We also evaluate leading MLLMs and find that they effectively localize events, integrate text and image inputs, and produce damage estimates aligned with seismic data. However, performance varies by language, modality, and prompt design, highlighting the need for further adaptation and robustness testing in real-world deployments. Our findings provide the first step toward globally scalable, cross-lingual disaster sensing with foundation models, and the released codes and prompts to support replication and future research.

\section*{Broader Impact and Ethics}
\subsection*{Broader Impact}
\paragraph*{Societal Relevance and Intended Use}
This work presents a scalable and multilingual framework for fine-grained earthquake damage assessment that leverages social media and MLLMs. By incorporating both textual and visual data, the system captures dimensions of disaster impact such as interior structural damage or personal distress, that are difficult to observe with conventional sensing systems. Our pipeline offers a lightweight, extensible tool for situational awareness, particularly in the early hours of a crisis when actionable information is limited. Through evaluations in Japan and the United States, we demonstrate the framework’s potential for global applicability. The methods and code are intended to be adaptable for other hazards (\emph{e.g.}, floods, wildfires) and use cases (\emph{e.g.}, infrastructure monitoring, rapid needs assessment).

\paragraph*{Inclusivity and Linguistic Diversity}
Disaster communication varies significantly across languages and cultures. Our framework is intentionally designed to support multilingual and multimodal inputs, allowing for more inclusive analysis across different user populations and platforms. The case study in Japan highlights the feasibility of applying foundation models beyond English-only settings, contributing to the growing body of work on equitable and linguistically diverse NLP applications. We encourage further development toward supporting low-resource languages and culturally grounded interpretations of crisis content.

\paragraph*{Interpretability and Human-AI Collaboration} 
We use prompting strategies such as chain-of-thought and few-shot examples to improve the transparency of multimodal model outputs. In addition to comparing predictions with official ground-truth seismic data (\emph{e.g.}, MMI levels), we include qualitative reasoning traces to assist with human interpretation. These steps enhance trust and traceability in model behavior, while positioning the system as a decision-support tool, not a replacement for expert review. This approach supports the responsible integration of LLMs into high-stakes environments like emergency management.

\subsection*{Ethics}
\paragraph*{Responsible Data Use and Privacy}
All data used in this study are drawn from publicly available social media posts, accessed via Twitter’s API under permitted use. Recognizing that disaster-related content is often shared under emotional duress, we employ several safeguards: no direct quotes or images are reproduced, user identifiers are removed, and results are reported only in aggregated spatial formats. Future deployments may benefit from further privacy-preserving measures such as differential privacy or on-device inference, particularly in operational settings.

\paragraph*{Robustness and Misinformation Risks}
Crisis-related social media can contain misinformation, rumors, or manipulated content. Our pipeline currently includes relevance filtering and heuristics for disaster-date alignment, but does not yet implement automated credibility detection. We view this as a key future direction, and recommend integration with source trustworthiness scoring and stance detection models for robust performance in noisy environments. These safeguards are particularly important in deployments where system outputs influence resource allocation or public messaging.

\paragraph*{Scope of Use and Deployment Guidance}
This pipeline is developed exclusively for public-interest applications such as disaster response, risk analysis, and resilience planning. It is not intended for use in surveillance, punitive actions, or insurance investigations. Responsible deployment requires human oversight, transparency about model limitations, and collaboration with emergency professionals and affected communities. We advocate for community-informed design and transparent documentation as this framework is adapted for real-world use.


\section{Appendices}

\label{sec:appendix}

\subsection{Damage-related filtering terms}
\begin{CJK}{UTF8}{min}
\begin{table*}[ht]
\centering
\caption{A list of terms used to filter ``damage-related'' tweets.}
\begin{tabular}{|c|p{12cm}|}
\hline
\textbf{Language} & \textbf{Damage-related words} \\
\hline
English & \small blackout, broke, broken, burn, burned, burning, burns, catastrophe, catastrophes, catastrophic, chaos, collapse, collapsed, collapses, crack, cracked, cracking, cracks, crash, crashed, crashes, cripple, cripples, crumble, crush, crushed, crushes, damage, damaged, damaging, dead, death, deaths, deform, deformed, deforms, demonish, destruct, destructed, destructing, destructs, destroy, destroyed, destroying, destroys, devastate, devastated, devastates, devastating, die, died, dies, displace, displaced, disrupt, disrupted, disrupting, disrupts, fatalities, fatality, fissure, fissures, fire, flood, flooded, flooding, hurt, hurting, hurts, injuries, injured, injury, kill, killed, killing, leak, leaked, leaking, leaks, massive, outage, rockslide, rubble, rupture, ruptures, safe, safety, scatter, scattered, scatters, severe, shatter, shattered, shatters, smash, smashed, smashes, smashing, suffer, suffered, suffering, suffers, trauma, warp, warps, wreck, wrecked, wrecks \\
\hline

Japanese & \small 停電(blackout，poweroutage), 壊れた(broke, broken), 燃える(burn), 燃えた(burned), 燃えている(burning), 大災害(catastrophe, catastrophes), 壊滅的(catastrophic), 混乱(chaos), 崩壊(collapse, collapsed, collapses), ひび(crack, cracked, cracking, cracks), 墜落(crash, crashed, crashes), 無力(cripple, cripples, helpless), 崩れる(crumble), 押しつぶす(crush, crushed, crushes), 損傷(damage, damaged, damaging), 死んだ(dead, died, die, dies), 死亡(death, deaths), 変形する(deform, deformed, deforms), 破壊(destruct, destructed, destructing, destructs), 破壊する(destroy, destroyed, destroying, destroys), 壊滅させる(devastate, devastated, devastates, devastating), 死ぬ(die, died, dies), 避難する(displace, displaced), 混乱する(disrupt, disrupted, disrupting, disrupts), 死者(fatalities, fatality), 裂け目(fissure, fissures), 火事(fire), 洪水(flood, flooded, flooding), 傷つく(hurt, hurting, hurts), けが(injuries, injury), 負傷した(injured)，殺す(kill, killed, killing), 漏れ(leak, leaked, leaking, leaks), 巨大な(massive), がけ崩れ(rockslide), 土砂崩れ(Landslide), 瓦礫(rubble), 破裂(rupture, ruptures), 安全(safe), 散らす(scatter, scattered, scatters), 厳しい(severe), 粉々にする(shatter, shattered, shatters), 打ち砕く(smash, smashed, smashes, smashing), 苦しむ(suffer, suffered, suffering, suffers), トラウマ(trauma), ゆがむ(warp, warps)   \\

\hline
\end{tabular}
\label{tab:damage_words}
\end{table*}
\end{CJK}

\subsection{MMI description}
\label{MMI}

\begin{table*}[htbp]
\centering
\small
\caption{MMI Intensity}
\resizebox{\textwidth}{!}{
\begin{tabular}{p{0.5cm} p{3cm} p{3cm} p{3cm} p{3cm}}
\toprule
\textbf{MMI} & \textbf{People's Reaction} & \textbf{Furnishings} & \textbf{Built Environment} & \textbf{Natural Environment} \\
\midrule
I & Not felt. &  &  & Changes in level and clarity of well water are occasionally associated with great earthquakes at distances beyond which the earthquakes felt by people. \\
\midrule
II & Felt by a few. & Delicately suspended objects may swing. &  & \\
\midrule
III & Felt by several; vibration like passing of truck. & Hanging objects may swing appreciably. &  & \\
\hline
IV & Felt by many; sensation like heavy body striking building. & Dishes rattle. & Walls creak; windows rattle. & \\
\midrule
V & Felt by nearly all; frightens a few. & Pictures swing out of place; small objects move; a few objects fall from shelves within the community. & A few instances of cracked plaster and cracked windows within the community. & Trees and bushes shaken noticeably. \\
\midrule
VI & Frightens many; people move unsteadily. & Many objects fall from shelves. & A few instances of fallen plaster, broken windows, and damaged chimneys within the community. & Some fall of tree limbs and tops, isolated rockfalls and landslides, and isolated liquefaction. \\
\midrule
VII & Frightens most; some lose balance. & Heavy furniture overturned. & Damage negligible in buildings of good design and construction, but considerable in some poorly built or badly designed structures; weak chimneys broken at roof line, fall of unbraced parapets. & Tree damage, rockfalls, landslides, and liquefaction are more severe and widespread with increasing intensity. \\
\midrule
VIII & Many find it difficult to stand. & Very heavy furniture moves conspicuously. & Damage slight in buildings designed to be earthquake resistant, but severe in some poorly built structures. Widespread fall of chimneys and monuments. & \\
\midrule
IX & Some forcibly thrown to the ground. &  & Damage considerable in some buildings designed to be earthquake resistant; buildings shift off foundations if not bolted to them. & \\
\midrule
X &  &  & Most ordinary masonry structures collapse; damage moderate to severe in many buildings designed to be earthquake resistant. & \\
\midrule
\end{tabular}
}
\end{table*}

\subsection{Model comparison}
\label{model_compartion}
Two annotators independently labeled a randomly selected sample of 50 tweets to evaluate inter-annotator reliability. We used Krippendorff’s Alpha ($\alpha$) (Equation \ref{eq:kripp_alpha}) to measure agreement, as it is a robust metric capable of handling multiple annotators, various data types (\emph{e.g.}, nominal, ordinal), and missing data \cite{artstein2017inter}. It also adjusts for chance agreement based on observed versus expected disagreement. The final alpha score was 0.67, indicating substantial agreement. This level of consistency is considered reasonable for subjective tasks involving nuanced, fine-grained classification.

\begin{equation}
\alpha = 1 - \frac{D_o}{D_e}
\label{eq:kripp_alpha}
\end{equation}

\noindent\textbf{Observed disagreement} \( D_o \) is calculated as:

\begin{equation}
D_o = \frac{1}{N} \sum_{i=1}^{N} \delta(a_{i1}, a_{i2})
\end{equation}

where:
\begin{itemize}
  \item \( N \) is the number of items,
  \item \( a_{i1}, a_{i2} \) are the annotations by two coders for item \( i \),
  \item \( \delta(a, b) = 1 \) if \( a \neq b \), and \( 0 \) if \( a = b \).
\end{itemize}

\noindent\textbf{Expected disagreement} \( D_e \) is computed from the marginal frequencies:

\begin{equation}
D_e = \sum_{c_1 \ne c_2} p(c_1) \cdot p(c_2)
\end{equation}

where:
\begin{itemize}
  \item \( p(c) = \frac{n_c}{2N} \) is the proportion of annotations assigned to category \( c \),
  \item \( n_c \) is the total number of times category \( c \) is used by both annotators,
  \item the denominator \( 2N \) is the total number of annotations across both coders.
\end{itemize}

\noindent\textbf{Interpretation}:
\begin{itemize}
  \item \( \alpha = 1 \): perfect agreement
  \item \( \alpha = 0 \): agreement equals chance
  \item \( \alpha < 0 \): worse than chance
\end{itemize}

\begin{table}[htbp]
\centering
\caption{Model comparison}
\resizebox{\columnwidth}{!}{  
\begin{tabular}{|l|c|c|c|}
\hline
\textbf{Model Name} & \textbf{Open Source} & \textbf{Accuracy} & \textbf{Price (\$)} \\
\hline
\textit{GPT-4.1} & No & 0.694 & 0.45 \\
\textit{GPT-4.1-mini} & No & 0.145 & 0.02 \\
\textit{GPT-4.1-nano} & No & -0.841 & 0.02 \\
\textit{GPT-4o} & No & 0.957 & 0.85 \\
\textit{GPT-4o-mini} & No & 0.706 & 0.15 \\
\textit{Gemini-2.5-Flash} & No & 0.775 & 0.15 \\
\textit{LLaVA 3-8B} & Yes & 0.113 & 0.00 \\
\textit{Qwen 2.5VL-7B} & Yes & 0.791 & 0.00 \\
\hline
\end{tabular}
}
\label{tab:model_comparison_extended}
\end{table}

To assess the cost-effectiveness of closed-source MLLMs, we monitored pricing across all eight evaluated models. Among them, Gemini-2.5-Flash was the most economically efficient and also demonstrated high alignment with human annotations. As a result, it was selected as the preferred closed-source multimodal model for our damage estimation tasks. For large-scale processing, we utilized the New York University High Performance Computing (NYU HPC) infrastructure, specifically the Greene cluster, which offers GPU-enabled nodes with NVIDIA Tesla V100 GPUs \cite{nyuhpc}. Within this environment, the complete analysis was executed in 2 to 3 days per event dataset.

\subsection{Prompt design}
\label{prompt design}
\begin{lstlisting}[language=Python]
LOCATION_PROMPT = """
Task:
You are a location identification expert. Your task is to determine whether a tweet is from a U.S.-based location, based on all available metadata and the tweet content.
Use the information below to infer the most granular geographic scale location if possible. Your output results must be generated after reasoning through  textual information.

Input:
Longitude: {longitude}
Latitude: {latitude}
Tweet Text: {tweet}
Location: {location}

Instruction:
Please follow the following identification steps
 Step 1: Check if Longitude, or Latitude exist. If so, infer the location and return it. Otherwise, move to Step 2.
 Step 2: Analyze the Tweet Text to find any explicit or implicit mention of a location (\emph{e.g.}, city, county, state, street, neighborhood, national park). If found, use it as the final location and return the most granular geographic information available. if not, move to step 3.
 Step 3: If neither one found in Step 1 and Step 2, use location fields from the input to infer location. 

Output Instructions:
If a U.S. location can be confidently identified, return it in plain text (\emph{e.g.}, "San Francisco, CA"). Avoid including non-physical locations (\emph{e.g.}, Earth, Galaxy).
If the tweet is not within the U.S. or the indeterminable, return "No".
If the tweet contains multiple locations, return the most granular geographic information.
If the final location information is abbreviated (\emph{e.g.}, "LV" for Las Vegas), return the full location name.
If the final location information contains distance information (\emph{e.g.}, "10 miles from LA"), or other vague details (\emph{e.g.}, "38th floor of hotel"), return "No".
Output must be in strict JSON format with the following structure:
{{
    "reasoning": "<Brief explanation of the reasoning steps taken>",
    "location": "<Provide final location information>"
}}
"""
\end{lstlisting}

\begin{lstlisting}[language=Python]
EVENT_PROMPT = """
Task: 
You are an earthquake engineer. Your task is to determine whether an input tweet is related to <2019 ridegcrest> earthquake in any meaningful way, such as their impact, damage, or aftermath.
Please read the tweet carefully and decide if it is about an earthquake. 

Input:
Tweet Text: {tweet}
    
Instruction:
Examples of tweets related to earthquakes:
-Last night she said that I needed to not stack all these shoe boxes up so high because an earthquake will happen and they will all fall on me! I am more worried about damaging the boxes and not being able to pass as Deadstock TBH than falling on me. 
-My outdoor pillows fell and my pancake is now burnt. This is the extent of the damage of the earthquake in Vegas for me. 
- Devi Bhujel, making tea in her kitchen in her village in Nepal. #water here is very hard. I take one jerrycan in a basket, it's about 10 liters maybe. The usual walking road is destroyed by the earthquake and construction. WaterAid/ Sibtain Haider #July4th 
Examples of tweets not related to earthquakes:
-we were watching CNN when they broke the news about the earthquake and the weather dude was like it "originated here" and circled the area near Tehachapi  which is where I'm going today and staying for the next couple days. 
-I knew those Trump tanks would cause damage.  #earthquake 

Restrictions: Exclude input tweet information if it solely contains magnitudes <\emph{e.g.},6.4 magnitudes>, distances from the epicenter <\emph{e.g.}, 10km> or other standard seismological data.  
Your output results must be generated after reasoning through  extual and/or visual information.

Output:
Respond only with Yes if the tweet is related to an earthquake.
Respond only with No if the tweet is not related to an earthquake.
Output must be in strict JSON format with the following structure:
{{
    "reasoning": "<Brief explanation of the reasoning steps taken>",
    "is_event_related": "<Yes | No>"
}}
"""
\end{lstlisting}

\begin{lstlisting}[language=Python]
IMAGE_ONLY_PROMPT:
f"""
Task: 
You are the earthquake damage assessment experts. Your task is to identify the damage level align with Modified Mercalli Intensity(MMI) levels from a given tweet. 
Your output must be generated based on evidence from the given tweet content.

Input:

Image Description:
Please analyze the image to assess the severity of the earthquake's damage. 

Instructions:

1. Human Impact Evaluation:
   Look for language or visual evidence suggesting that people experienced or emotionally reacted to the earthquake. Indicators may include expressions or signs of: fear(\emph{e.g.}, "people were terrified", "panic in the streets"), shock or confusion(\emph{e.g.}, "people didn't know what to do"), physical presence or impact (\emph{e.g.}, "people ran outside", "rescue teams helping trapped residents"), sensation reporting (\emph{e.g.}, "I felt the floor shake", "it was the strongest I've ever felt"), etc. Then return:
    1: if there is any mention or evidence of human emotional or physical experience of the earthquake. 
    0: if there is no indication that humans were present or affected emotionally/physically.

2. Damage Type Classification:
   Classify the damage type as either:
    Interior: Damage that is clearly observed inside a building (e,g, cracked or collapsed interior walls, broken windows or glass, displaced or fallen indoor furniture, ceiling or floor cracks, shaking fixtures (\emph{e.g.}, light fixtures, shelves)).
    Exterior: Damage that is clearly observed on the outside of buildings or in the surrounding environment (\emph{e.g.}, Collapsed buildings, shifts in building foundation or roof collapse, partial structural failure, cracked roads/sidewalks/bridges, fallen trees or utility poles, visible debris or rubble outside).
    Both: Evidence of damage is present both inside and outside of structures. The content includes clear indicators of both categories listed above.
    None: The input does not provide enough information to determine whether the damage is interior, exterior, or both.

3. Damage Level Classification (MMI Scale):
   After identifying the damage type (Interior, Exterior, Both, or None) and human impact ("1" or "o"), classify the earthquake damage level align with MMI scale.
   If human impact is 1 from the previous step (human can feel the earthquake), consider both human impact and damage level classification.
   If human impact is 0 from the previous step (human can't feel the earthquake), proceed based solely with damage level classification.
 
   Damage Level Categories (MMI Scale):
   1 - Not felt: No noticeable damage.
   2 - Weak: Felt by only a few people at rest; no damage to buildings.
   3 - Light: Felt indoors, especially on upper floors; no significant structural damage.
   4 - Moderate: Felt by most people; some damage to buildings, such as minor cracks.
   5 - Strong: Felt by everyone; damage to buildings, minor cracks, but no collapse.
   6 - Very Strong: Damage to buildings, visible structural deformation.
   7 - Severe: Significant damage, some collapses or structural failures.
   8 - Very Severe: Many buildings collapse or are severely damaged.
   9 - Violent: Total destruction in some areas, severe damage.
   10 - Extreme: Complete destruction of all structures in the affected area.


Output:
Output must be in strict JSON format with the following structure:
{{
    "human_impact": <1 or 0>,
    "damage_type": "<Interior | Exterior | Both | None>",
    "damage_level": <1-10>,
    "reasoning": "<Explain how you get the human_impact, damage_type, damage_level based on the input information>",
    "confidence": "<Return how confident (scale 0-1) you are in the final MMI damage level>"
}}
"""
    
TEXT_IMAGE_FUSION_PROMPT:
f"""
Task: 
You are the earthquake damage accessment experts. Your task is to identify the damage level align with Modified Mercalli Intensity(MMI) levels from a given tweet. 
Your output must be generated based on evidence from the given tweet content.

Input: 
Text Description:
{tweet}

Image Description:
Please analyze the image to assess the severity of the earthquake's damage based on MMI Scale. 

Instructions:

1. Human Impact Evaluation:
   Look for language or visual evidence suggesting that people experienced or emotionally reacted to the earthquake. Indicators may include expressions or signs of: fear(\emph{e.g.}, "people were terrified", "panic in the streets"), shock or confusion(\emph{e.g.}, "people didn't know what to do"), physical presence or impact (\emph{e.g.}, "people ran outside", "rescue teams helping trapped residents"), sensation reporting (\emph{e.g.}, "I felt the floor shake", "it was the strongest I've ever felt"), etc. Then return:
    1: if there is any mention or evidence of human emotional or physical experience of the earthquake. 
    0: if there is no indication that humans were present or affected emotionally/physically.

2. Damage Type Classification:
   Classify the damage type as either:
    - Interior: Damage that is clearly observed inside a building (e,g, cracked or collapsed interior walls, broken windows or glass, displaced or fallen indoor furniture, ceiling or floor cracks, shaking fixtures (\emph{e.g.}, light fixtures, shelves)).
    - Exterior: Damage that is clearly observed on the outside of buildings or in the surrounding environment (\emph{e.g.}, Collapsed buildings, shifts in building foundation or roof collapse, partial structural failure, cracked roads/sidewalks/bridges, fallen trees or utility poles, visible debris or rubble outside).
    - Both: Evidence of damage is present both inside and outside of structures. The content includes clear indicators of both categories listed above.
    - None: The input does not provide enough information to determine whether the damage is interior, exterior, or both.

3. Damage Level Classification (MMI Scale):
   After identifying the damage type (Interior, Exterior, Both, or None) and human impact ("1" or "o"), classify the earthquake damage level align with MMI scale.
   If human impact is 1 from the previous step (human can feel the earthquake), consider both human impact and damage level classification.
   If human impact is 0 from the previous step (human can't feel the earthquake), proceed based solely with damage level classification.
 
   Damage Level Categories (MMI Scale):
   1 - Not felt: No noticeable damage.
   2 - Weak: Felt by only a few people at rest; no damage to buildings.
   3 - Light: Felt indoors, especially on upper floors; no significant structural damage.
   4 - Moderate: Felt by most people; some damage to buildings, such as minor cracks.
   5 - Strong: Felt by everyone; damage to buildings, minor cracks, but no collapse.
   6 - Very Strong: Damage to buildings, visible structural deformation.
   7 - Severe: Significant damage, some collapses or structural failures.
   8 - Very Severe: Many buildings collapse or are severely damaged.
   9 - Violent: Total destruction in some areas, severe damage.
   10 - Extreme: Complete destruction of all structures in the affected area.


Output:
Output must be in strict JSON format with the following structure:
{{
    "human_impact": <1 or 0>,
    "damage_type": "<Interior | Exterior | Both | None>",
    "damage_level": <1-10>,
    "reasoning": "<Explain how you get the human_impact, damage_type, damage_level based on the input information>",
    "confidence": "<Return how confident (scale 0-1) you are in the final MMI damage level>"
}}
"""
\end{lstlisting}

\subsection{Example of 3M pipeline outputs}
\label{prompt_output}
The following table \ref{tab:tweet_model_results} presents a representative example of how the three selected MLLMs—LLaVA, Qwen, and Gemini—analyze a tweet containing both text and image information. All three models accurately identify the location (El Monte, CA) and confirm the tweet’s event relevance. While their MMI level estimates and confidence scores are similar, the models differ slightly in how they classify damage type (interior vs. exterior) and assess human impact. The reasoning outputs provide further insight into each model’s interpretive process, revealing how text and image inputs are integrated to support the final prediction. This example highlights the overall consistency of model outputs while also illustrating subtle differences in how damage is inferred from multimodal content. 

\renewcommand{\arraystretch}{1.5}

\begin{table*}[htbp]
\centering
\small
\caption{Comparison of model responses to a tweet example.}
\label{tab:tweet_model_results}
\begin{tabular}{|p{0.22\textwidth}|>{\centering\arraybackslash}p{0.18\textwidth}|>{\centering\arraybackslash}p{0.14\textwidth}|>{\centering\arraybackslash}p{0.14\textwidth}|>{\centering\arraybackslash}p{0.14\textwidth}|}

\hline
\textbf{Tweet Example} & \textbf{Model Responses} & \textbf{LLaVA} & \textbf{Qwen} & \textbf{Gemini} \\
\hline
\multirow{7}{=}{\includegraphics[width=\linewidth, height=3.5cm, keepaspectratio]{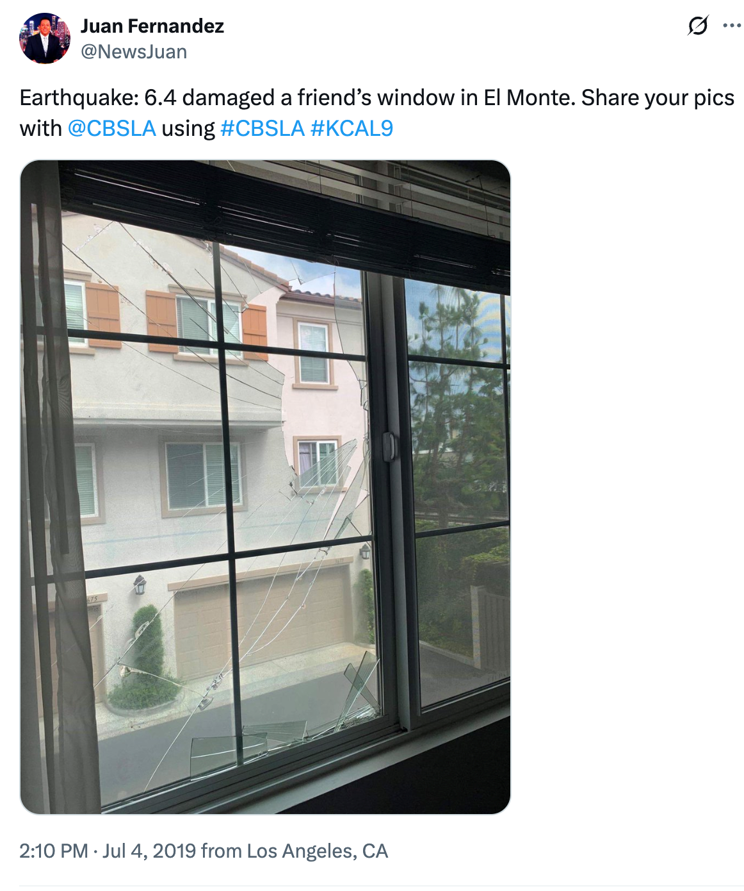}}
& Location Response & El Monte, CA & El Monte, CA & El Monte, CA \\
\cline{2-5}
& Event Response & Yes & Yes & Yes \\
\cline{2-5}
& Human-impacted & No & No & Yes \\
\cline{2-5}
& Damage Type & Exterior & Interior & Interior \\
\cline{2-5}
& MMI levels & 3 & 4 & 4 \\
\cline{2-5}
& Voted candidate & Text and Image & Text and Image & Text and Image \\
\cline{2-5}
& Confidence & 0.8 & 0.9 & 0.9 \\
\hline
{\textbf{Reasoning for \textit{LLaVA}}} & \multicolumn{4}{p{0.66\textwidth}|}{The tweet describes a 6.4 earthquake that damaged a friend's window in El Monte. The image shows a window with visible damage, but there is no indication of human impact or damage to the interior of the building. The damage is classified as exterior damage, and the damage level is classified as MMI 3, which corresponds to a light earthquake felt indoors, especially on upper floors; no significant structural damage} \\
\hline
\textbf{Reasoning for \textit{Qwen}} & \multicolumn{4}{p{0.66\textwidth}|}{The image shows a broken window, indicating interior damage. The text does not mention any human emotional or physical reaction to the earthquake, so the human impact is considered 0. The broken window suggests moderate damage, which aligns with the MMI scale category of 'Moderate' (4)} \\
\hline
\textbf{Reasoning for \textit{Gemini}} & \multicolumn{4}{p{0.66\textwidth}|}{Both the text and image contribute to the assessment. The text indicates that a window was damaged, implying human impact as the earthquake was felt. The image confirms that the damage is to the interior of the building, specifically a broken window. This level of damage aligns with MMI level 4, which includes minor damage to buildings such as broken windows or minor cracks. The combined evidence gives a more reliable assessment} \\
\bottomrule

\end{tabular}
\end{table*}

\subsection{Integrated maps}
\label{maps}
To better understand the spatial distribution and alignment of model-predicted damage levels, we present integrated visualizations of the MLLM outputs overlaid with official ground-truth MMI contours. These maps allow for intuitive comparison between predicted damage intensity and observed seismic impacts, offering insight into each model’s spatial coverage, localization precision, and capacity for fine-grained damage differentiation. The visualizations highlight both the geographic extent and spatial resolution of the models’ estimation, facilitating a qualitative assessment of model performance across different regions.

\begin{figure*}[htbp]
    \centering
    \includegraphics[trim=0 160 0 100, clip, width=\textwidth]{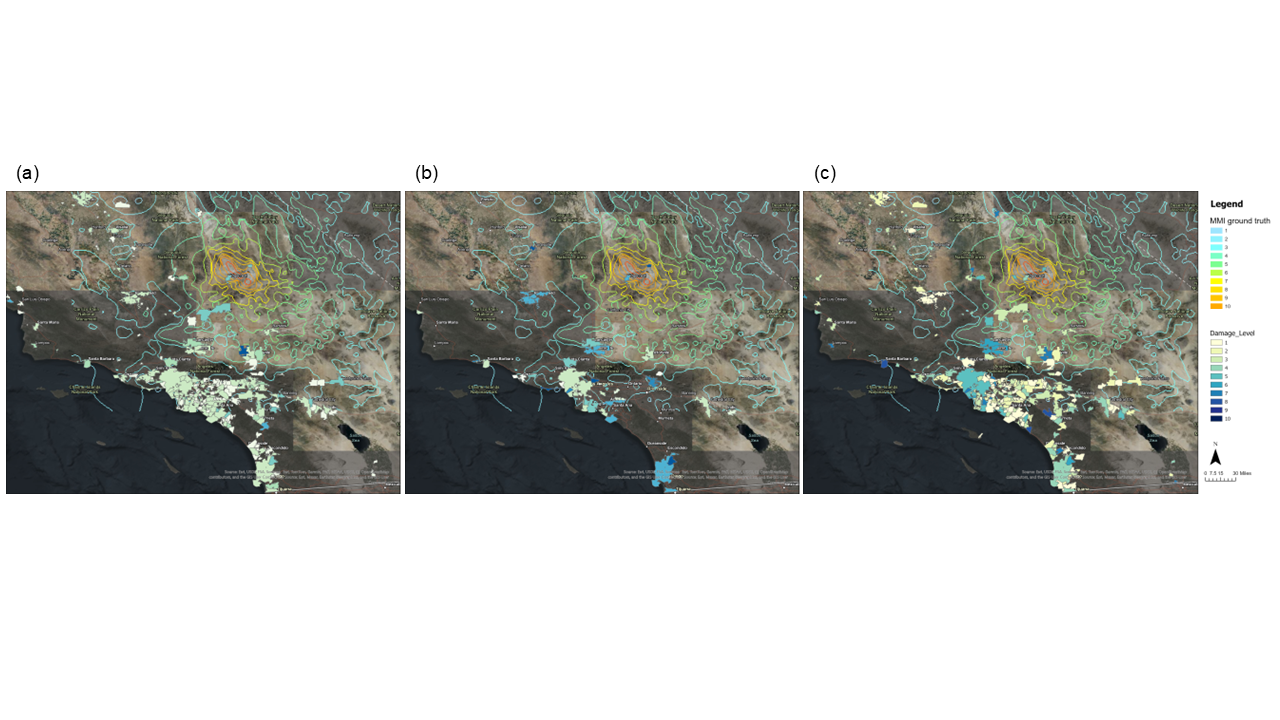}
    \caption{Integrated map for the 2019 Ridgecrest earthquake from (a) LLaVA 3-8B, (b) Qwen-2.5-VL-7B, and (c) Gemini-2.5-Flash}
    \vspace{2mm}
    \includegraphics[trim=0 160 0 100, clip, width=\textwidth]{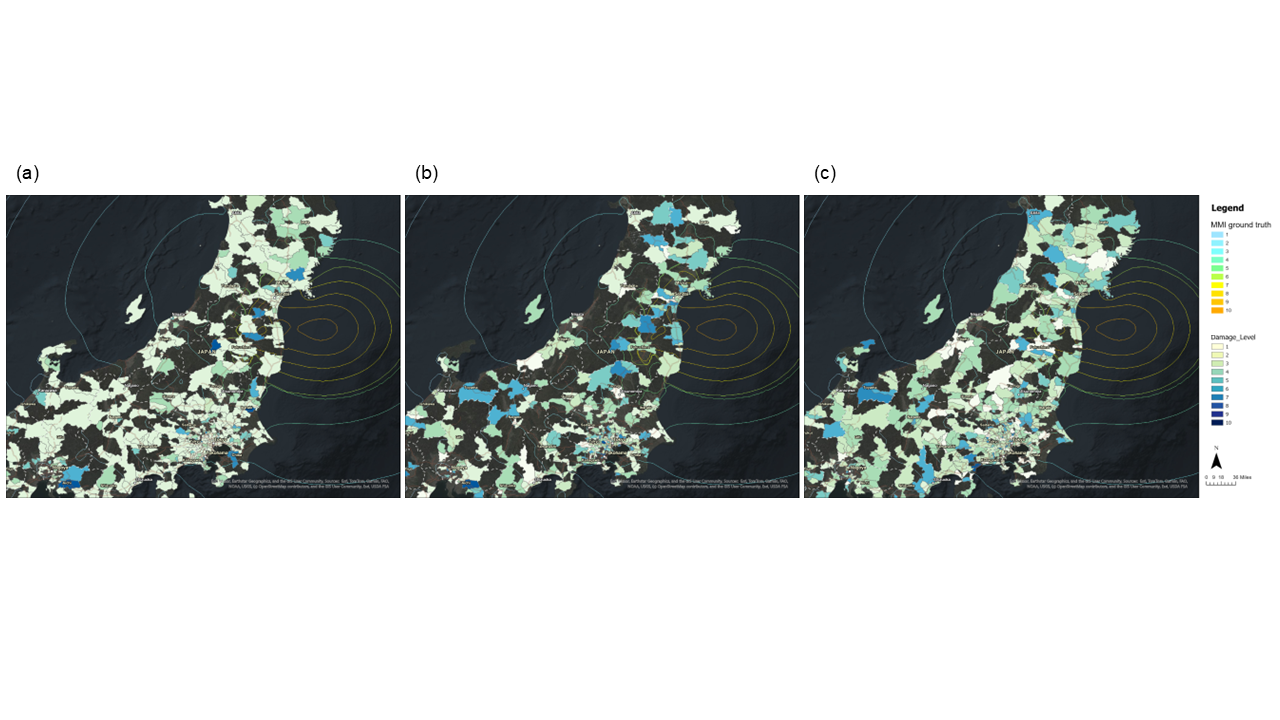}
    \caption{Integrated map for the 2022 Fukushima earthquake from (a) LLaVA 3-8B, (b) Qwen-2.5-VL-7B, and (c) Gemini-2.5-Flash}
    \label{fig: Map}
\end{figure*}

\subsection{Prompt rewritten versions}
\label{prompt_versions}
\begin{lstlisting}[language=Python]
PROMPT_V2: """
Task:
You are an earthquake damage assessment expert. For each tweet, follow these three steps to classify the damage:

Step 1: Describe any human emotional or physical reactions mentioned in the tweet or shown in the image.
Step 2: Describe any structural or environmental damage observed in the image.
Step 3: Based on both observations, classify the earthquake's Modified Mercalli Intensity (MMI) level.

Input:
Text Description:
{tweet}

Image Description:
Please analyze the image to assess visible earthquake damage.

Output:
Respond in JSON format:
{{
    "human_impact": <1 or 0>,
    "damage_type": "<Interior | Exterior | Both | None>",
    "damage_level": <1-10>,
    "reasoning": "<Step-by-step breakdown>",
    "confidence": "<0.0-1.0>"
}}
"""
\end{lstlisting} 

\begin{lstlisting}[language=Python]
PROMPT_V3: f"""
Task:
Your primary role is to assess earthquake damage using visual cues in the image provided. Use the tweet text only if needed to resolve ambiguities.

Input:
Image Description:
Analyze for any visible earthquake damage-structural collapse, debris, road cracks, etc.

Text Description:
{tweet}

Output:
Return the damage classification in JSON:
{{
    "human_impact": <1 or 0>,
    "damage_type": "<Interior | Exterior | Both | None>",
    "damage_level": <1-10>,
    "reasoning": "<Visual evidence used to support the output>",
    "confidence": "<0.0-1.0>"
}}
"""
\end{lstlisting} 

\begin{lstlisting}[language=Python]
PROMPT_4: f"""
Analyze the tweet and associated image to determine the earthquake damage level according to the MMI scale.

Input:
Text: {tweet}
Image: [image provided]

Output:
Strictly return JSON:
{{
    "human_impact": <1 or 0>,
    "damage_type": "<Interior | Exterior | Both | None>",
    "damage_level": <1-10>,
    "reasoning": "<Why each field was chosen>",
    "confidence": "<0.0-1.0>"
}}
"""
\end{lstlisting} 

\begin{lstlisting}[language=Python]
PROMPT_5: f"""
Task:
Please answer the following questions based on the tweet and image:

1. Did people seem to experience or react to the earthquake?
2. Where did the damage occur-inside, outside, both, or unclear?
3. What is the MMI level based on the human and structural impact?

Tweet: {tweet}
Image: [Analyze the image]

Output:
Output must be in strict JSON format with the following structure:
{{
    "human_impact": <1 or 0>,
    "damage_type": "<Interior | Exterior | Both | None>",
    "damage_level": <1-10>
    "reasoning": "<Explain how you get the human_impact, damage_type, damage_level based on the input information>",
    "confidence": "<Return how confident (scale 0-1) you are in the final MMI damage level>"
}}
"""
\end{lstlisting} 

\begin{lstlisting}[language=Python]
PROMPT_6: f"""
Task:
Review the following examples and then analyze the new tweet and image.

Example 1:
Tweet: "People ran outside screaming after their house walls cracked."
Image: [shows rubble and collapsed roof]
Output:
{{
    "human_impact": 1,
    "damage_type": "Both",
    "damage_level": 7,
    "reasoning": "Clear human fear and both interior (walls) and exterior (roof) damage.",
    "confidence": "0.85"
}}

Now classify:
Tweet: {tweet}
Image: [Analyze the image]

Output:
Output must be in strict JSON format with the following structure:
{{
    "human_impact": <1 or 0>,
    "damage_type": "<Interior | Exterior | Both | None>",
    "damage_level": <1-10>
    "reasoning": "<Explain how you get the human_impact, damage_type, damage_level based on the input information>",
    "confidence": "<Return how confident (scale 0-1) you are in the final MMI damage level>"
}}
"""
\end{lstlisting} 

\begin{lstlisting}[language=Python]
PROMPT_7: f"""
Task:
Classify the tweet and image below according to the following strict schema.

Input:
Tweet Content: {tweet}
Image Content: [image provided]

Output Format:
All fields must match format:
- human_impact: (0 or 1)
- damage_type: "Interior", "Exterior", "Both", or "None"
- damage_level: Integer from 1 to 10
- reasoning: Text, <400 characters
- confidence: Float between 0 and 1

Output:
Output must be in strict JSON format with the following structure:
{{
    "human_impact": <1 or 0>,
    "damage_type": "<Interior | Exterior | Both | None>",
    "damage_level": <1-10>,
    "reasoning": "<Explain how you get the human_impact, damage_type, damage_level based on the input information>",
    "confidence": "<Return how confident (scale 0-1) you are in the final MMI damage level>"
}}
"""
\end{lstlisting} 

\subsection{Satellite image vs MLLMs results}

Remote sensing offers critical supplementary insights into the environmental repercussions of seismic events. Building upon our prior reasoning analysis, we conduct an in-depth evaluation of environmental impacts, with a particular emphasis on damage typologies, to further assess the model's inferential capabilities.

 We acquire remote sensing data through the utilization of the Scene Classification Map (SCM) derived from Sentinel-2 Level-2A products\cite{Sentinel2}. The SCM is generated via the Sen2Cor processor, which implements a series of threshold-based assessments on top-of-atmosphere reflectance data across multiple spectral bands to categorize each pixel into predefined classes, including vegetation, water, soil/desert, snow, clouds, and shadows\cite{Aybar2022}. This classification facilitates the differentiation of land cover types and the detection of alterations attributable to seismic disturbances. The SCM is available at spatial resolutions of 20 m and 60 m and encompasses quality indicators for cloud and snow probabilities\cite{Jelének2021}.


 The correlation between tweets referencing exterior damage and the ground-truth MMI values is presented in  Table \ref{tab:ex_cor_JP} and Table \ref{tab:ex_cor_CA}. We first observe that Gemini demonstrates a stronger capability in identifying exterior damage impacts, as it detected the highest number of relevant tweets. Additionally, Gemini achieved relatively higher performance in estimating exterior damage severity, showing weak to moderate positive Pearson correlation scores, whereas the other two models exhibited only weak correlations. These findings suggest that future research should consider integrating remote sensing data, particularly for assessing damage to natural and built environments, to further enhance the accuracy of MMI scale estimation.

\begin{table*}[htbp]
\centering
\small
\caption{Fukushima, Japan Validation Result between MMI Ground Truth(Extorior)}
\begin{tabular}{lcccccc}
\toprule
&
\multicolumn{2}{c}{\textbf{Gemini}} &
\multicolumn{2}{c}{\textbf{Qwen}} &
\multicolumn{2}{c}{\textbf{LLaVA}} \\
\midrule

Pearson- R& \multicolumn{2}{c}{0.54}& \multicolumn{2}{c}{0.05}& \multicolumn{2}{c}{0.12}\\
Number of Tweets& \multicolumn{2}{c}{1207}& \multicolumn{2}{c}{86}& \multicolumn{2}{c}{24}\\
\bottomrule
\end{tabular}
\label{tab:ex_cor_JP}
\end{table*}

\begin{table*}[htbp]
\centering
\small
\caption{Ridgecrest, CA Validation Result between MMI Ground Truth(Exterior)}
\begin{tabular}{lcccccc}
\toprule
&
\multicolumn{2}{c}{\textbf{Gemini}} &
\multicolumn{2}{c}{\textbf{Qwen}} &
\multicolumn{2}{c}{\textbf{LLaVA}} \\
\midrule

Pearson- R& \multicolumn{2}{c}{0.38}& \multicolumn{2}{c}{0.09}& \multicolumn{2}{c}{0.26}\\
Number of Tweets& \multicolumn{2}{c}{1185}& \multicolumn{2}{c}{264}& \multicolumn{2}{c}{53}\\
\bottomrule
\end{tabular}
\label{tab:ex_cor_CA}
\end{table*}

\section*{Limitations}

While this study provides a scalable and generalizable pipeline for multimodal earthquake damage assessment, it has several limitations that should be considered when interpreting the results. First, the use of social media introduces inherent sampling biases. Prior studies have shown that Twitter users are disproportionately younger, more educated, urban, and male, which limits the demographic representativeness of the data \cite{pew2022twitter}. This population bias can reduce the generalizability of findings, particularly in contexts where equitable disaster response is critical. Moreover, disparities in internet access and digital infrastructure further constrain data coverage. Global digital divides and infrastructural disruptions in disaster-affected regions may result in missing or delayed social media signals. These conditions reduce the utility of social media as a ground-level information source during large-scale disasters. 

Second, the data retrieval process itself imposes restrictions. In our study, tweets were collected using a single keyword (“earthquake”) and filtered using a manually defined set of damage-related terms. While this approach provides a focused dataset, it may miss relevant posts that use alternative vocabulary or regional expressions. Consequently, reliance on fixed keyword libraries can limit recall and introduce topic filtering bias, especially across languages and local dialects.
Second, the data retrieval process itself imposes restrictions. In our study, tweets were collected using a single keyword (“earthquake”) and filtered using a manually defined set of damage-related terms. While this approach provides a focused dataset, it may miss relevant posts that use alternative vocabulary or regional expressions. Consequently, reliance on fixed keyword libraries can limit recall and introduce topic filtering bias, especially across languages and local dialects.

Third, our study employs foundation MLLMs without task-specific fine-tuning. While our approach highlights the models’ general capabilities, fine-tuning on domain-specific or multilingual disaster corpora could improve prediction accuracy, robustness, and contextual alignment. 

Fourth, we limited our full-scale evaluation to three selected models (from an initial pool of eight) based on a balance of performance and computational cost. This choice reflects practical deployment considerations, especially for real-time use in embodied agents. However, further exploration with larger or instruction-tuned models may yield different performance dynamics and should be explored in future work.

Finally, the use of human-reported DYFI data as ground truth introduces subjectivity and potential inconsistencies. 
These crowd-sourced labels, while widely adopted in earthquake research, are subjective and may vary due to perceptual or reporting biases. 
Incorporating additional data sources, such as structural damage assessments, seismic sensor data, or building inspection records, could provide a more comprehensive benchmark for future evaluations.


\bibliography{custom}

\appendix

\end{document}